\pdfoutput=1
\documentclass[11pt]{article}
\usepackage{acl}
\usepackage{times}
\usepackage{latexsym}
\usepackage{booktabs}
\usepackage{multirow}
\usepackage{tabularx}
\usepackage{graphicx}
\usepackage{amsmath}
\usepackage[T1]{fontenc}
\usepackage[utf8]{inputenc}
\usepackage{inconsolata}
\usepackage{adjustbox}
\usepackage{array}

\newcolumntype{R}[2]{%
  >{\adjustbox{angle=#1,lap=\width-(#2)}\bgroup}%
  l%
  <{\egroup}%
}
\newcommand*\rot{\multicolumn{1}{R{90}{-1em}}}
\usepackage{microtype}
\usepackage{listings}
\usepackage{xcolor}
\colorlet{punct}{red!60!black}
\definecolor{background}{HTML}{EEEEEE}
\definecolor{delim}{RGB}{20,105,176}
\colorlet{numb}{magenta!60!black}
\lstdefinelanguage{json}{
  basicstyle=\normalfont\ttfamily,
  numbers=left,
  numberstyle=\scriptsize,
  stepnumber=1,
  numbersep=8pt,
  showstringspaces=false,
  breaklines=true,
  frame=lines,
  backgroundcolor=\color{background},
  literate=
  *{:}{{{\color{punct}{:}}}}{1}
  {,}{{{\color{punct}{,}}}}{1}
  {\{}{{{\color{delim}{\{}}}}{1}
  {\}}{{{\color{delim}{\}}}}}{1}
  {[}{{{\color{delim}{[}}}}{1}
  {]}{{{\color{delim}{]}}}}{1},
}
\usepackage{xspace}
\newcommand{\opspam}{\textsc{OpSpam}\xspace}
\newcommand{\crosscult}{\textsc{CrossCultDe}\xspace}
\newcommand{\dereva}{\textsc{DeRev2014}\xspace}
\newcommand{\derevb}{\textsc{DeRev2018}\xspace}
\newcommand{\derevab}{\textsc{DeRev2014/2018}\xspace}
\newcommand{\opendomain}{\textsc{OpenDomain}\xspace}
\newcommand{\trial}{\textsc{Trial}\xspace}
\newcommand{\bltc}{\textsc{BLTC}\xspace}
\newcommand{\onlinede}{\textsc{OnlineDe}\xspace}
\newcommand{\mafiascum}{\textsc{Mafiascum}\xspace}
\newcommand{\mud}{\textsc{MU3D}\xspace}
\newcommand{\bol}{\textsc{BoxOfLies}\xspace}
\newcommand{\diplomacy}{\textsc{Diplomacy}\xspace}
\newcommand{\decop}{\textsc{DecOp}\xspace}
\newcommand{\bluff}{\textsc{Bluff}\xspace}
\newcommand{\corpusname}{\textsc{UniDecor}\xspace}
\newcommand{\F}{$\text{F}_1$\xspace}

\usepackage{annotates}

\title{\corpusname:\\ A \underline{Uni}fied \underline{De}ception \underline{Cor}pus for Cross-Corpus Deception Detection}

\author{%
  Aswathy Velutharambath$^{1,2}$ \and Roman Klinger$^{1}$\\
  $^1$Institut f\"ur Maschinelle Sprachverarbeitung, University of
  Stuttgart, Germany \\
  $^2$100 Worte Sprachanalyse GmbH, Heilbronn, Germany\\
  \texttt{aswathy.velutharambath@100Worte.de}\\
  \texttt{roman.klinger@ims.uni-stuttgart.de}\\
}

\begin{document}
\maketitle

\begin{abstract}
  Verbal deception has been studied in psychology, forensics, and
  computational linguistics for a variety of reasons, like
  understanding behaviour patterns, identifying false testimonies, and
  detecting deception in online communication. Varying motivations
  across research fields lead to differences in the domain choices to
  study and in the conceptualization of deception, making it hard 
  to compare models and build robust deception
  detection systems for a given language. With this
  paper, we improve this situation by surveying available
  English deception datasets which include domains like social media
  reviews, court testimonials, opinion statements on specific topics,
  and deceptive dialogues from online strategy games. We consolidate
  these datasets into a single unified corpus. Based on this resource,
  we conduct a correlation analysis of linguistic cues of deception
  across datasets to understand the differences and perform
  cross-corpus modeling experiments which show that a cross-domain
  generalization is challenging to achieve. The unified deception 
  corpus (\corpusname) can be obtained from 
  \url{https://www.ims.uni-stuttgart.de/data/unidecor}.

\end{abstract}

\section{Introduction}
Deception detection has remained an area of vested interest in fields
like psychology, forensics, law, and computational linguistics for a
myriad of reasons like understanding behavioral patterns of lying 
\cite{Newman2003, depaulo_morris_2004},
identifying fabricated information \cite{fakenews}, 
distinguishing false statements or
testimonies \cite{trial} and detecting deception in online communication
\cite{digital-deception-hancock}.
These are relevant tasks because of
the truth bias, which is the inherent inclination of humans to
actively believe or passively presume that a statement made by another
person is true and accurate by default, without the need for evidence to
substantiate this belief \cite{truth-bias}. While this facilitates
efficient communication, it also makes people susceptible to
deception, especially in online media where digital deception
\cite{digital-deception-hancock} manifests in many forms like fake
news, misleading advertisements, impersonation and scams. This
warrants automatic deception detection systems that can accurately
distinguish between truthful and deceptive discourse solely from
textual data.

The task of automatic deception detection comes with several challenges.
Deception or lying is a complex human behavior and its signals are
faint in text. Moreover, it is sensitive to the communication context,
interlocutors, and the stake involved \cite{ten2012cry, 
salvetti-etal-2016-tangled}. Most
importantly, acquiring annotated data proves to be one of the major
hurdles for deception studies. Traditional data annotation methods cannot
be employed because human performance is shown to be
worse than machines in differentiating truths and lies
\cite{depaulo-2006, Vrij}. One way to collect accurate data is to get
 the labels at source by the person producing the text. Alternatively, 
 they can be collected using the acquired knowledge that certain types 
 of contents are deceptive.  Across the literature, different strategies
  like crawling fake reviews \cite{yao-etal-2017-online}, collecting
text from users identified as suspicious \cite{fornaciari2020fake},
using non-linguistic deception cues
\cite{fornaciari-poesio-2014-identifying} and soliciting through
crowd-sourcing \cite{ott-etal-2011-finding, ott-etal-2013-negative,
  salvetti-etal-2016-tangled} have been employed to obtain reliable
annotations.

The diversity in the domains of interest, the medium of deceptive
communication (spoken vs.\ written) and dataset creation procedures
make it difficult to compare cues of deception across datasets and to
understand their generalizability across different domains.  With this
study, we aim at mitigating this situation by conducting a comparative
survey of publicly available textual deception datasets.  We
contribute (1) a consolidated corpus in a unified format and (2)
conduct experiments in which we evaluate models trained on one data
set on all others. Our (3) results show that cross-corpus,
particularly cross-domain, generalizability is limited, which
motivates future work to develop robust deception detectors. We lay
the foundation for such work with (4) additional correlation analyses
of the linguistic cues of deception across datasets and verify their
generalizability across domains.

\section{Background \& Related Work}

Deception in communication is the act of intentionally causing another
person to have a false belief that the deceiver knows or believes to
be false \citep{zuckerman-definition1981, mahon2007definition,
  digital-deception-hancock}.  Lies, exaggerations, omissions, and
distortions are all different forms of deception \cite{turner-75,
  metts-89}. While the definition of deception varies across
literature, they concur that it is intentional or deliberate
\cite{mahon2007definition, gupta2012telling}.

\subsection{Domains and Ground Truth}  
Deception research is spread across disciplines which contributed to a
variety of domains and consequentially to a number of data collection
methods. Domains include opinions statements on a specific topic
\cite{perez-rosas-mihalcea-2014-cross,capuozzo-etal-2020-decop, mu3d},
open domain statements \cite{perez-rosas-mihalcea-2015-experiments},
online reviews \cite{ott-etal-2011-finding, ott-etal-2013-negative,
  fornaciari-poesio-2014-identifying, yao-etal-2017-online}, deceptive
dialogues in strategic games like
Mafiascum\footnote{\url{https://www.mafiascum.net/}}, Box of Lies and
Diplomacy \citep{Ruiter2018TheMD, soldner-etal-2019-box,
  peskov-etal-2020-takes, Skalicky2020PleasePJ} and court trials
\cite{trial}.

The ground truth generation strategies differ across datasets.  
While datasets of opinion statements
\cite{perez-rosas-mihalcea-2014-cross,capuozzo-etal-2020-decop, mu3d},
and online reviews \cite{ott-etal-2011-finding,
  ott-etal-2013-negative, fornaciari-poesio-2014-identifying,
  yao-etal-2017-online} are collected in written form, interviews
include both verbal and non-verbal content \cite{trial}. Game-based 
corpora contain monologue \cite{Skalicky2020PleasePJ} or dialogue 
data \cite{soldner-etal-2019-box, peskov-etal-2020-takes}.

All of these resources contain instances that are labeled as truthful
or deceptive. Only few studies employ the same procedure to generate
both truthful and deceptive content \citep{salvetti-etal-2016-tangled,
  Skalicky2020PleasePJ}; most resort to separate strategies for
collecting them \cite{ott-etal-2011-finding, ott-etal-2013-negative,
  fornaciari2020fake}.  Instances labeled as deceptive are either
solicited content or collected from a source identified as
deceptive. \citet{ott-etal-2011-finding,ott-etal-2013-negative}
crawled the truthful reviews from websites of interest and the
deceptive ones were crowd-sourced through AMT\footnote{Amazon’s
  Mechanical Turk, \label{amt}\url{https://www.mturk.com/}}, while
\citet{salvetti-etal-2016-tangled} solicited both via AMT.
\citet{yao-etal-2017-online} tracked fake review generation tasks
from crowd-sourcing platforms to identify deceptive reviews and
reviewers. For the datasets based on strategic games, the labels are
assigned based on game rules. Opinion domain datasets contain stances
on topics, like gay marriage and abortion, written by the same person,
where the truthful labeled opinions align with the author's true
opinion and deceptive ones align with the opposite
\citep{perez-rosas-mihalcea-2014-cross,capuozzo-etal-2020-decop}.

\subsection{Automatic Deception Detection Methods}
Several studies have explored the effectiveness of automatic methods
to detect deception from textual data. These include feature-based
classification methods with support vector machines
\citep{ott-etal-2011-finding,perez-rosas-mihalcea-2014-cross,
  fornaciari-poesio-2014-identifying}, logistic regression
\cite{Ruiter2018TheMD}, decision trees
\cite{perez-rosas-mihalcea-2015-experiments}, and random forests
\cite{soldner-etal-2019-box,
  perez-rosas-mihalcea-2015-experiments}. Some studies also consider
contextual information by using recurrent neural networks
\cite{peskov-etal-2020-takes} and transformer-based models
\citep{capuozzo-etal-2020-decop, peskov-etal-2020-takes,
  fornaciari-etal-2021-bertective}. Transformers are not always better
-- \citet{peskov-etal-2020-takes} show that BERT is en par with LSTMs
while \citet{fornaciari-etal-2021-bertective} showed that adding extra
attention layers help to improve upon the previous state of the art.

Most works focused on modeling the concept of deception in one
domain. An exception is \citet{HernandezCastaneda2016} who report
cross-domain classification results on \opspam, \dereva, and
\crosscult, but in an all-against-one setting, not in a
one-against-one setup.

\subsection{Linguistic Cues of Deception}
\label{sec:cues}
To understand the phenomenon of deception better, 
previous studies have analyzed the linguistic cues that 
characterize deceptive language in written statements, 
spoken conversations, and online communication
\citep{Newman2003,Bond2005} and demonstrated that a systematic
analysis of these cues can prove valuable in automated deception
detection specifically in computer-mediated communication
\cite{Zhou2004}. \citet{Newman2003} noted that the use of fewer 
self-references in deceptive statements indicate that the liars are
attempting to distance themselves from the lies. The use of exclusive
words (e.g., \textit{but}, \textit{rather}) allow deceivers to
introduce communicative ambiguity into the discourse. \citet{hancock}
noted that these cues are broadly associated with the number of words,
use of pronouns, use of emotion words, and presence of markers of
cognitive complexity. They also pointed out that these cues can
manifest differently based on the type and medium of discourse;
real-world vs.\ online or monologue vs.\ dialogue. 

While these analyses have found application in
machine learning models, there are more sets of features that have
been used to automatically detect deception. These include n-grams
\citep{fornaciari-poesio-2014-identifying, fornaciari2020fake,
  ott-etal-2011-finding}, part-of-speech tags \citep{mu3d,
  fornaciari2020fake, perez-rosas-mihalcea-2015-experiments},
lexicon-based features, including the Linguistic Inquiry and Word
Count \citep[LIWC,][]{pennebaker-2015-linguistic} psychological categories,
\citep{perez-rosas-mihalcea-2014-cross, yao-etal-2017-online} and
production rules derived from syntactic context free grammar trees
\cite{yao-etal-2017-online,perez-rosas-mihalcea-2015-experiments}.
\citet{duran_hall_mccarthy_mcnamara_2010}, \citet{pinocchio-2012} and
\citet{linguistic-cues-meta-2015} conducted extensive surveys and
analyses of different linguistic cues of deception.

\section{Unified Deception Dataset}
As preparation for cross-corpus analysis of the concept of
deception, we consolidate publicly available textual deception
datasets into a unified format.\footnote{We refer to our corpus as
  \corpusname: ``Unified Deception Corpus''. The scripts to download
  and convert the dataset can be found in the following repository: \url{https://www.ims.uni-stuttgart.de/data/unidecor}} We now describe the
included datasets.

\begin{table*}
  \centering\small
  \setlength{\tabcolsep}{3.5pt}

\begin{tabular}{llrrrrr}
	\toprule

\multicolumn{1}{l}{Dataset} & \multicolumn{1}{l}{Domain} & \multicolumn{1}{l}{Truthful} & \multicolumn{1}{l}{Deceptive} & \multicolumn{1}{l}{Total} & \multicolumn{1}{l}{TC} & \multicolumn{1}{l}{SC} \\
\cmidrule(r){1-1}\cmidrule(r){2-2}\cmidrule(r){3-3}\cmidrule(r){4-4}\cmidrule(r){5-5}\cmidrule(r){6-6}\cmidrule{7-7}
	Bluff the listener (\bluff) &      game &    251 (33.3\%) &   502 (66.7\%) &    753 &   241.66 &   11.5 \\
	Diplomacy dataset (\diplomacy) &      game &  16402 (94.9\%) &    887 (\phantom{0}5.1\%) &  17289 &    24.53 &    1.7 \\
	Mafiascum dataset (\mafiascum) &      game &   7439 (76.9\%) &  2237 (23.1\%) &   9676 &  4690.69 &  362.8 \\
	Multimodal Decep. in Dialogues (\bol) &      game &    101 (20.2\%) &   400 (79.8\%) &    501 &     12.2 &    1.6 \\
	Miami University Decep. Detection Db. (\mud) & interview &    160 (50.0\%) &   160 (50.0\%) &    320 &    131.7 &    5.7 \\
	Real-life trial data (\trial) & interview &     60 (49.6\%) &    61 (50.4\%) &    121 &    79.85 &    3.9 \\
	Cross-cultural deception (\crosscult) &   opinion &    600 (50.0\%) &   600 (50.0\%) &   1200 &     80.0 &    4.5 \\
	Deceptive Opinion (\decop) &   opinion &   1250 (50.0\%) &  1250 (50.0\%) &   2500 &    65.56 &    4.0 \\
	Boulder Lies and Truth Corpus (\bltc) &    review &   1041 (69.8\%) &   451 (30.2\%) &   1492 &   116.92 &    6.5 \\
	Deception in reviews (\dereva) &    review &    118 (50.0\%) &   118 (50.0\%) &    236 &   145.22 &    6.7 \\
	Deception in reviews (\derevb) &    review &   1552 (50.0\%) &  1552 (50.0\%) &   3104 &    176.6 &    8.1 \\
	Deceptive opinion spam (\opspam) &    review &    800 (50.0\%) &   800 (50.0\%) &   1600 &    170.5 &    9.5 \\
	Online deceptive reviews (\onlinede) &    review & 101431 (85.9\%) & 16694 (14.1\%) & 118125 &    171.5 &    7.2 \\
	Open Domain Deception (\opendomain) & statement &   3584 (50.0\%) &  3584 (50.0\%) &   7168 &     9.33 &    1.0 \\
\midrule
	 &      & 134789 (82.1\%) & 29296 (17.9\%) & 164085 &   436.88 &  31.05 \\
	\bottomrule
\end{tabular}
  \caption{Datasets included in our unified corpus (\corpusname),
    together with statistical information. TC: average token count;
    SC: average sentence count.}
  \label{tab:distribution}
\end{table*}

\newcommand{\corpusnamehead}[1]{\noindent\textbf{#1}}

\corpusnamehead{Deceptive Opinion Spam (\opspam).}
\citet{ott-etal-2011-finding} describes \textit{deceptive opinion
  spam} as fraudulent reviews written to sound authentic with the goal
to deceive the reader. To study the nature of such reviews, they
collected truthful reviews by crawling online review platforms like
TripAdvisor\footnote{\url{https://www.tripadvisor.com/}} and
crowd-sourced deceptive reviews via Amazon’s Mechanical Turk
(AMT). The initial \opspam dataset published by
\citet{ott-etal-2011-finding} contains 400 truthful and 400 deceptive
reviews with positive sentiments. \citet{ott-etal-2013-negative}
extended the dataset to include reviews with negative sentiments. The
complete \opspam dataset contains 1600 instances labeled for veracity
and sentiment. It is available publicly with a Creative Commons
Attribution-NonCommercial-ShareAlike
license.\footnote{\url{https://myleott.com/op-spam.html}}

\corpusnamehead{Cross-cultural Deception (\crosscult).}
\citet{perez-rosas-mihalcea-2014-cross} collected the \crosscult
dataset to investigate deception in a cross-cultural setting. It
consists of short essays on the topics of abortion, death penalty, and
feelings about a best friend, collected from the United States, India, and
Mexico. We take into account the data collected from the United States
and India which are in English and consist of 100 deceptive and 100
truthful essays per topic per geographical region adding up to 1200
labeled instances. The dataset is available for download without
mentioning any usage
restrictions.\footnote{\label{rada}\url{https://web.eecs.umich.edu/~mihalcea/downloads.html}}

\corpusnamehead{Deception in Reviews (\derevab).}  To investigate the
phenomenon of sock puppetry,
\citet{fornaciari-poesio-2014-identifying} collected \dereva,
containing book reviews from \textit{amazon.com} that were identified
as authentic or fake using predefined linguistic cues.  To overcome
the shortcoming that these cues cannot be used while developing a
deception classifier, \citet{fornaciari2020fake} released the \derevb
dataset, in which they collect deceptive reviews based on \textit{a
  priori} knowledge about authors who solicited fake reviews.
Additionally, the authors crowd-sourced both truthful and deceptive reviews 
for the same books.  The \dereva dataset contains 118 reviews each with a
truthful label and a deceptive label, while the \derevb dataset
includes 1552 reviews each collected from \textit{amazon.com} and
through crowd-sourcing with a balanced distribution of truthful and
deceptive reviews. The datasets overlap by 62 reviews. Both corpora
are available for
download.\footnote{\url{https://fornaciari.netlify.app/}}
 
\corpusnamehead{Open Domain Deception (\opendomain).}
\citet{perez-rosas-mihalcea-2015-experiments} study deception, gender,
and age detection with an open domain dataset acquired via AMT. Workers
were asked to contribute seven true and seven plausible deceptive
statements without a restriction of domain, each in a single sentence.
The balanced dataset consists of 7168 annotated instances with
additional demographic information. The data set is made available 
without specifying usage restrictions.\textsuperscript{\ref{rada}}

\corpusnamehead{Real-life Trial Data (\trial).} To study real-life
high-stake deception scenarios, \citet{10.1145/2818346.2820758}
collected videos of trial hearings from publicly available sources
like ``The Innocence Project'' 
website\footnote{\url{http://www.innocenceproject.org/}}. The dataset
contains multimodal information with annotations for non-verbal
behavior like facial displays and gestures in addition to
crowd-sourced transcriptions. It contains 60 truthful and 61
deceptive reviews.This corpus is made available without 
specifying any usage
restrictions.\textsuperscript{\ref{rada}}

\corpusnamehead{Boulder Lies and Truth Corpus (\bltc).}
\citet{salvetti-etal-2016-tangled} built a balanced dataset containing
reviews elicited via AMT for the domains of electronic appliances and
hotels. The crowd-workers were instructed to write fake or real
reviews, with positive or negative sentiment, about objects that they
were familiar with or not. Unlike other datasets which limited the
labeling to truthful vs.\ deceptive, this dataset distinguished
between fake and deceptive reviews, where the former are fabricated
opinions about an unknown object while the latter was a false review
of a known object. The corpus contains 1492 reviews, out of which 451
are truthful and the rest is labeled as fake or deceptive. It
is available through the LDC.\footnote{Linguistic Data
Consortium, \url{https://catalog.ldc.upenn.edu/LDC2014T24}}

\corpusnamehead{Online Deceptive Reviews (\onlinede).} To address the
bottleneck that large realistic data for deception detection do
not exist, \citet{yao-etal-2017-online} created the \onlinede corpus
containing manipulated reviews posted online.  They employed the
automatic deception detection framework outlined by \citet{fazayi2015}
to identify deceptive reviewers and reviews from social media
manipulation campaigns. It contains more than 100K labeled
reviews with $\approx$10000 deceptive instances, covering more than 30
domains. The dataset is available for research purposes from the
authors.

\corpusnamehead{Mafiascum Dataset (\mafiascum).} This dataset
published by \citet{Ruiter2018TheMD} contains a collection of more
than 700 games of Mafia, an online strategy game played on the
Internet forum
\mafiascum\footnote{\url{https://www.mafiascum.net/}}. Here, players
are assigned deceptive or non-deceptive roles randomly, which serve
as annotations of the instances. Each of the 9000 documents contain 
all messages written by a single user in a specific
game. The average token count in the instances ($4690.69$) is
therefore considerably higher than in other corpora. The authors have
made the dataset publicly available along with the code used for
analyses.\footnote{\url{https://bitbucket.org/bopjesvla/thesis/src/master/}}

\corpusnamehead{Miami University Deception Detection Database (\mud).}
To investigate the role of gender and race in deception studies,
\citet{mu3d} created \mud. It is a collection of interview videos
where participants were instructed to talk truthfully or deceptively
about their relationship with a person whom they liked or disliked.
The 80 participants, each belonging to a different gender and ethical
background contributed to a positive truth, a negative truth, a
positive lie and a negative lie, counting to 160 truthful and 160
deceptive interview content.  The transcriptions of these videos along
with demographic information, valency, and veracity annotations are
made available for research
purposes with a Creative Commons Attribution-NonCommercial-NoDerivs license.\footnote{\url{https://sc.lib.miamioh.edu/handle/2374.MIA/6067}}

\corpusnamehead{Multimodal Deception in Dialogues (\bol).}  To explore
deception in conversational dialogue, \citet{soldner-etal-2019-box}
collected the \bol dataset which is based on the ``Box of Lies'' game,
a segment on ``The Tonight Show Starring Jimmy Fallon'' where two
celebrity guests take turns describing the contents of a box but are
allowed to lie. The opposing player must decide if they believe the
description or not.  The collected dataset contained 25 videos of the
game, transcribed and annotated for non-verbal cues of deception and
the veracity of the describer. We exported the statements containing 
veracity label from the dataset using 
ELAN\footnote{\url{https://tla.mpi.nl/tools/tla-tools/elan/download/}}, 
a tool used to create and modify annotations for audio and video data. 
The dataset is available for download without specifying any usage
restrictions.\textsuperscript{\ref{rada}}

\corpusnamehead{Diplomacy Dataset (\diplomacy).}  To study deception
in a conversational context specifically in long-lasting
relationships, \citet{peskov-etal-2020-takes} employed the
negotiation-based online game \diplomacy.  The players use deception
as a strategy to convince other players to form alliances, for which
they use a chat interface. Contrary to other deception datasets,
\diplomacy contains an additional label for perceived truthfulness of
an instance.  The intended and perceived truthfulness of each message
was annotated by the sender and the receiver respectively. Out of more
than 13k messages less than 5\% are labeled as intended or perceived
lie, resulting in an imbalanced dataset. We use the dataset made
available through
ConvoKit.\footnote{\url{https://convokit.cornell.edu/documentation/diplomacy.html}}

\corpusnamehead{Deceptive Opinion (\decop).} To study deception in
multi-domain and multi-lingual settings,
\citet{capuozzo-etal-2020-decop}, following the method described by
\citet{perez-rosas-mihalcea-2014-cross}, collected truthful and
deceptive opinion statements on five different topics, namely
abortion, cannabis legalization, euthanasia, gay marriage, and
policies on migrants.  The experiment was conducted for English and
Italian, from which we include the English instances in
\corpusname. They consist of 2500 opinions statements with balanced
labels. This dataset can be obtained from the authors.

\corpusnamehead{Bluff the Listener (\bluff).} To study humorous
deception with no malicious intent, \citet{Skalicky2020PleasePJ}
compiled the \bluff dataset. It contains data from the ``Bluff the
Listener'' game which is part of the radio show ``Wait\ldots Don’t
Tell Me''.  It is a variation of the game ``Two Truths and a
  Lie'' in which a panelist tells three stories, two of which are
true, and one of which is false.  This corpus published by
\citet{Skalicky2020PleasePJ} contains 753 humorous stories collected
from 251 episodes broadcast from 2010 to 2019. The authors downloaded
the transcripts from News-Bank\footnote{\url{www.newsbank.com}}, a
curated repository containing current and archived media. One-third of
the stories are truthful while two-thirds are fabricated, counting to
251 truthful and 502 deceptive stories. The dataset is publicly
available and can be downloaded via the OSF
platform.\footnote{\url{https://osf.io/download/mupd9}}

\paragraph{Aggregation.} We consolidate the datasets into one unified
corpus in which each instance is assigned a binary label indicating if
it is truthful or deceptive. We retain annotation dimensions that are
available for more than one dataset (age, gender, country, and
sentiment). More details on the aggregation process and a sample entry
from the corpus are available in
Appendix~\ref{append:aggregation}. Table~\ref{tab:distribution}
provides an overview of the corpora, including size, label
distribution, token and sentence counts\footnote{Using NLTK's
  wordpunct\_tokenize and sent\_tokenize}, along with the domain. The
datasets vary greatly in its size, but the distribution of labels is
mostly comparable, except for \bltc, \onlinede and \diplomacy with
comparably high counts for truthful instances.

\begin{figure*}
  \centering
  \includegraphics[scale=0.5]{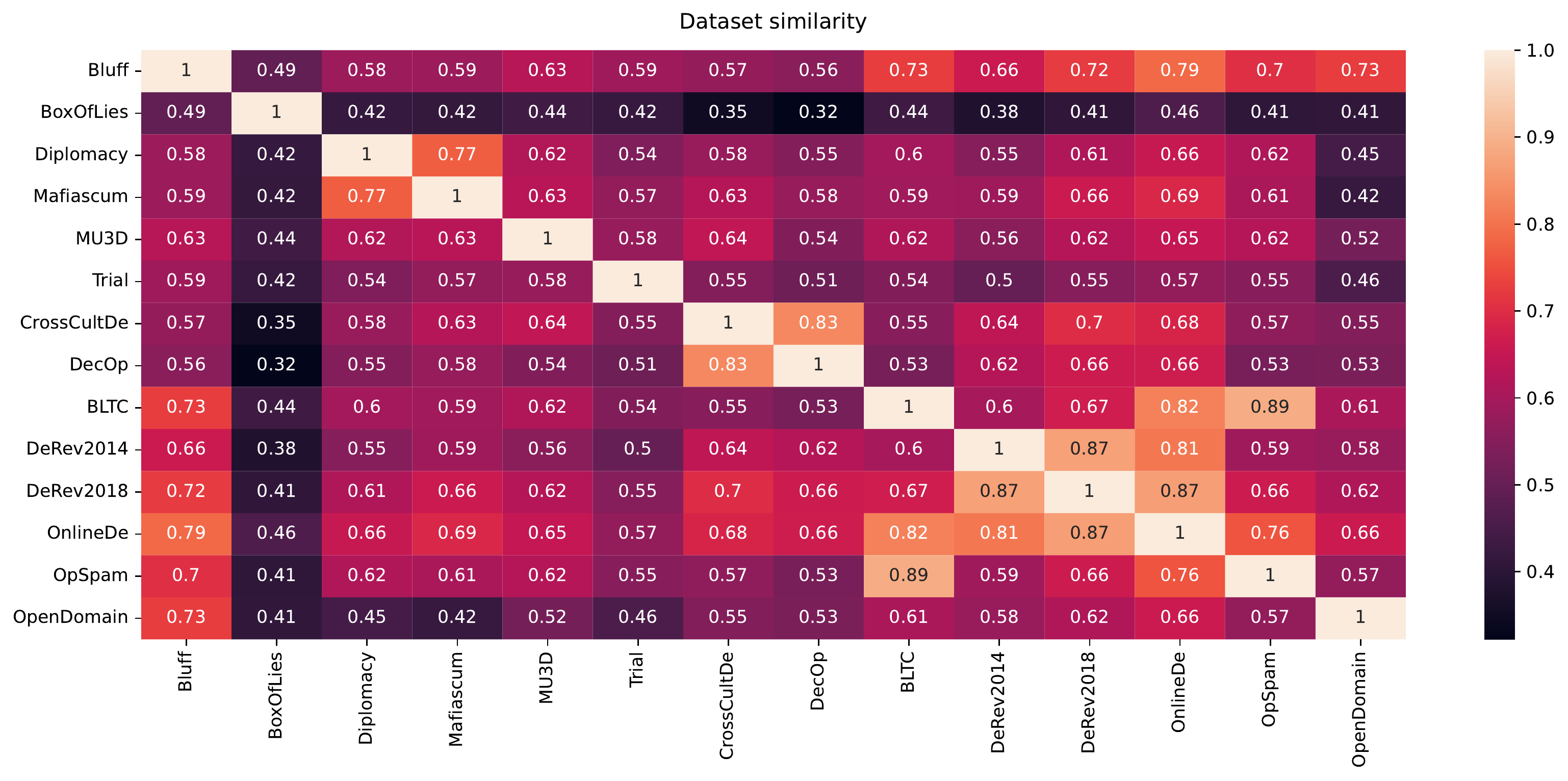}
  \caption{Similarity values, according to the measure 
    proposed by \citet{LI2022103377}, between all pairs of datasets.}
  \label{fig:source-heatmap}
\end{figure*}

\section{Similarity Analysis}
\label{ss}

The datasets included in \corpusname come from a variety of domains
and differ markedly in terms of the method of collection. At the same
time, datasets from the same domains also have differences (e.g.,
solicited reviews vs.\ actual reviews). To understand the differences
of datasets better, we explore the similarity values between these
datasets using the corpus-similarity measure defined by
\citet{LI2022103377}, which uses word unigram frequencies and
character trigram frequencies of the datasets to calculate the
Spearman’s $\rho\in[-1;1]$.\footnote{We use the Python implementation
  \url{https://github.com/jonathandunn/corpus\_similarity}}

Figure \ref{fig:source-heatmap} shows a symmetrical matrix of
similarity scores for dataset pairs. The correlation values could in
principle be negative, but we do not observe any such values because
all corpora are in the same language and have a high
degree of term and character frequency overlap.

The heatmap reflects the domains of datasets. For instance, \bltc,
\opspam, as well as \dereva and \derevb from the review domain have
similarity scores of $0.89$ and $.87$, respectively.  The opinion
statement datasets \crosscult and \decop exhibit a high similarity
score of $0.83$.  Similarly, \mafiascum and \diplomacy show relatively
high similarity ($0.77$), despite differences in the game rules.
 
Datasets obtained under specific conditions within the same domain are
assigned a lower similarity score. For instance, \bol, which is a game
that takes place in an in-person setting, differs from the online game
datasets ($.42$ with \diplomacy and \mafiascum). We also observe
similarity across domains, e.g., \bluff is more similar to reviews
than games, presumably due to its monologue setting instead of
dialogue.

\begin{table*}
  \small
  \setlength{\tabcolsep}{4pt}
  \begin{tabular}{lrrrrrrrrrrrrrr}
    \toprule
    &   \multicolumn{14}{c}{Datasets} \\
    \cmidrule{2-15}
    Features &  \rot{\bltc} &  \rot{\bluff} &  \rot{\bol} &  \rot{\crosscult} &  \rot{\decop} &  \rot{\dereva} &  \rot{\derevb} &  \rot{\diplomacy} &  \rot{\mafiascum} &   \rot{\mud} &  \rot{\onlinede} &  \rot{\opendomain} &  \rot{\opspam} &  \rot{\trial} \\
    \midrule
    Analytic        &    $\mathbf{.13}$ &   $-.04$ &     $\mathbf{.12}$ &        $.01$ &    $.02$ &    $\mathbf{-.25}$ &     $\mathbf{.23}$ &        $\mathbf{.02}$ &  $\mathbf{-.02}$ &   $\mathbf{.14}$ &    $\mathbf{.10}$ &     $\mathbf{.05}$ &   $\mathbf{.15}$ &   $\mathbf{.25}$ \\
    Authentic       &    $.03$ &   $-.05$ &      $.00$ &       $\mathbf{.28}$ &   $\mathbf{.22}$ &     $\mathbf{.28}$ &    $\mathbf{-.05}$ &        $\mathbf{-.03}$ &   $-.02$ &    $.07$ &     $.00$ &    $\mathbf{-.04}$ &  $\mathbf{-.09}$ &   $-.09$ \\
    BigWords        &    $.02$ &     $.00$ &    $\mathbf{.18}$ &        $.04$ &   $\mathbf{.05}$ &    $\mathbf{-.21}$ &     $\mathbf{.24}$ &          $.01$ &   $-.01$ &   $\mathbf{.18}$ &  $\mathbf{-.01}$ &     $\mathbf{.03}$ & $\mathbf{-.08}$ &    $.09$ \\
    Clout           &    $.00$ &     $.00$ &      $.02$ &      $\mathbf{-.11}$ &  $\mathbf{-.28}$ &    $\mathbf{-.45}$ &      $.00$ &         $\mathbf{.02}$ &   $\mathbf{.02}$ &    $.03$ &  $\mathbf{-.05}$ &      $.01$ &    $\mathbf{.10}$ &   $\mathbf{.26}$ \\
    Cognition       &  $\mathbf{-.08}$ &   $\mathbf{.17}$ &     $-.05$ &        $.02$ &   $\mathbf{.07}$ &     $-.06$ &    $\mathbf{-.13}$ &         $-.01$ &   $-.01$ &  $\mathbf{-.17}$ &     $.00$ &    $\mathbf{-.09}$ &  $\mathbf{-.06}$ &  $\mathbf{-.28}$ \\
    GunningFog &   $\mathbf{.18}$ &  $\mathbf{-.21}$ &     $\mathbf{.12}$ &       $\mathbf{.21}$ &   $\mathbf{.25}$ &      $.01$ &     $\mathbf{.13}$ &        $\mathbf{-.09}$ &  $\mathbf{-.03}$ &   $-.04$ &   $\mathbf{.13}$ &      $.02$ &    $.02$ &    $.06$ \\
    Kincaid         &   $\mathbf{.18}$ &  $\mathbf{-.21}$ &     $\mathbf{.14}$ &        $\mathbf{.2}$ &   $\mathbf{.24}$ &      $.01$ &     $\mathbf{.13}$ &        $\mathbf{-.08}$ &  $\mathbf{-.03}$ &   $-.04$ &   $\mathbf{.13}$ &     $\mathbf{.03}$ &    $.02$ &    $.06$ \\
    Linguistic      &  $\mathbf{-.07}$ &    $\mathbf{.10}$ &    $\mathbf{-.15}$ &        $.04$ &    $\mathbf{.10}$ &     $\mathbf{.29}$ &    $\mathbf{-.14}$ &        $\mathbf{-.02}$ &  $\mathbf{-.03}$ &  $\mathbf{-.16}$ &  $\mathbf{-.05}$ &    $\mathbf{-.05}$ &  $\mathbf{-.18}$ &   $-.08$ \\
    Period          &    $.01$ &  $\mathbf{-.07}$ &      $.02$ &      $\mathbf{-.11}$ &  $\mathbf{-.18}$ &     $\mathbf{.26}$ &    $\mathbf{-.07}$ &          $.00$ &     $.00$ &    $.03$ &   $\mathbf{.01}$ &     $\mathbf{.03}$ &   $\mathbf{.24}$ &   $-.06$ \\
    Physical        &    $.02$ &    $.03$ &     $\mathbf{.15}$ &       $-.04$ &  $\mathbf{-.16}$ &    $\mathbf{-.25}$ &     $\mathbf{.06}$ &          $.00$ &   $\mathbf{.03}$ &    $.04$ &  $\mathbf{-.15}$ &     $-.01$ &   $-.01$ &    $.06$ \\
    WC              &   $\mathbf{.18}$ &  $\mathbf{-.21}$ &      $.04$ &      $\mathbf{.22}$ &  $\mathbf{.25}$ &      $.02$ &     $\mathbf{.13}$ &         $\mathbf{-.10}$ &    $.01$ &   $-.04$ &   $\mathbf{.13}$ &     $-.02$ &    $.02$ &    $.06$ \\
    auxverb         &  $\mathbf{-.08}$ &   $\mathbf{.12}$ &     $-.06$ &      $\mathbf{-.08}$ &  $\mathbf{-.09}$ &     $\mathbf{.22}$ &    $\mathbf{-.12}$ &         $-.01$ &   $\mathbf{.02}$ &  $\mathbf{-.15}$ &    $.00$ &     $\mathbf{.03}$ &  $\mathbf{-.08}$ &  $\mathbf{-.21}$ \\
    focusfuture     &  $\mathbf{-.09}$ &   $\mathbf{.09}$ &     $-.02$ &       $-.04$ &  $\mathbf{-.08}$ &    $\mathbf{-.17}$ &     $\mathbf{-.2}$ &         $-.01$ &    $.02$ &   $-.04$ &   $\mathbf{.01}$ &    $\mathbf{-.04}$ &  $\mathbf{-.16}$ &    $.08$ \\
    function        &   $-.05$ &   $\mathbf{.13}$ &     $-.03$ &         $.00$ &    $\mathbf{.10}$ &     $\mathbf{.25}$ &    $\mathbf{-.06}$ &        $\mathbf{-.04}$ &  $\mathbf{-.03}$ &  $\mathbf{-.15}$ &  $\mathbf{-.03}$ &    $\mathbf{-.05}$ &  $\mathbf{-.23}$ &  $\mathbf{-.23}$ \\
    i               &  $\mathbf{-.06}$ &  $\mathbf{-.15}$ &     $-.07$ &       $\mathbf{.13}$ &    $\mathbf{-.3}$ &     $\mathbf{.39}$ &    $\mathbf{-.16}$ &        $\mathbf{-.05}$ &    $.02$ &   $-.01$ &  $\mathbf{-.12}$ &    $\mathbf{-.04}$ &  $\mathbf{-.33}$ &   $-.13$ \\
    shehe           &    $.01$ &  $\mathbf{-.11}$ &     $-.03$ &      $\mathbf{-.15}$ &    $.00$ &    $\mathbf{-.17}$ &    $\mathbf{-.07}$ &           $.00$ &  $\mathbf{-.04}$ &  $\mathbf{-.14}$ &  $\mathbf{.04}$ &    $\mathbf{-.04}$ &   $-.01$ &  $\mathbf{-.18}$ \\
    verb            &  $\mathbf{-.11}$ &    $.07$ &    $\mathbf{-.09}$ &      $\mathbf{-.06}$ &  $\mathbf{-.07}$ &     $\mathbf{.16}$ &    $\mathbf{-.26}$ &        $\mathbf{-.02}$ &     $.00$ &  $\mathbf{-.14}$ &  $\mathbf{-.07}$ &     $-.01$ &  $\mathbf{-.16}$ &   $-.14$ \\
    you             &   $\mathbf{-.10}$ &   $\mathbf{.17}$ &     $-.03$ &       $-.05$ &  $\mathbf{-.07}$ &    $\mathbf{-.19}$ &    $\mathbf{-.23}$ &          $.01$ &   $\mathbf{.03}$ &   $-.08$ &  $\mathbf{-.05}$ &    $\mathbf{-.05}$ &    $.01$ &   $-.05$ \\
    \bottomrule
  \end{tabular}
  \caption{Point-biserial correlation between the deception
    labels and linguistic features (LIWC categories + readability). We
    only show features with a correlation coefficient of $\geq .15$  and
    $p\leq .05$ for at least three datasets. Correlation scores with
    $p\leq .05$ are shown in bold.}
  \label{tab:linguistic}
\end{table*}

\begin{figure*}[t]
  \centering
  \includegraphics[scale=0.5]{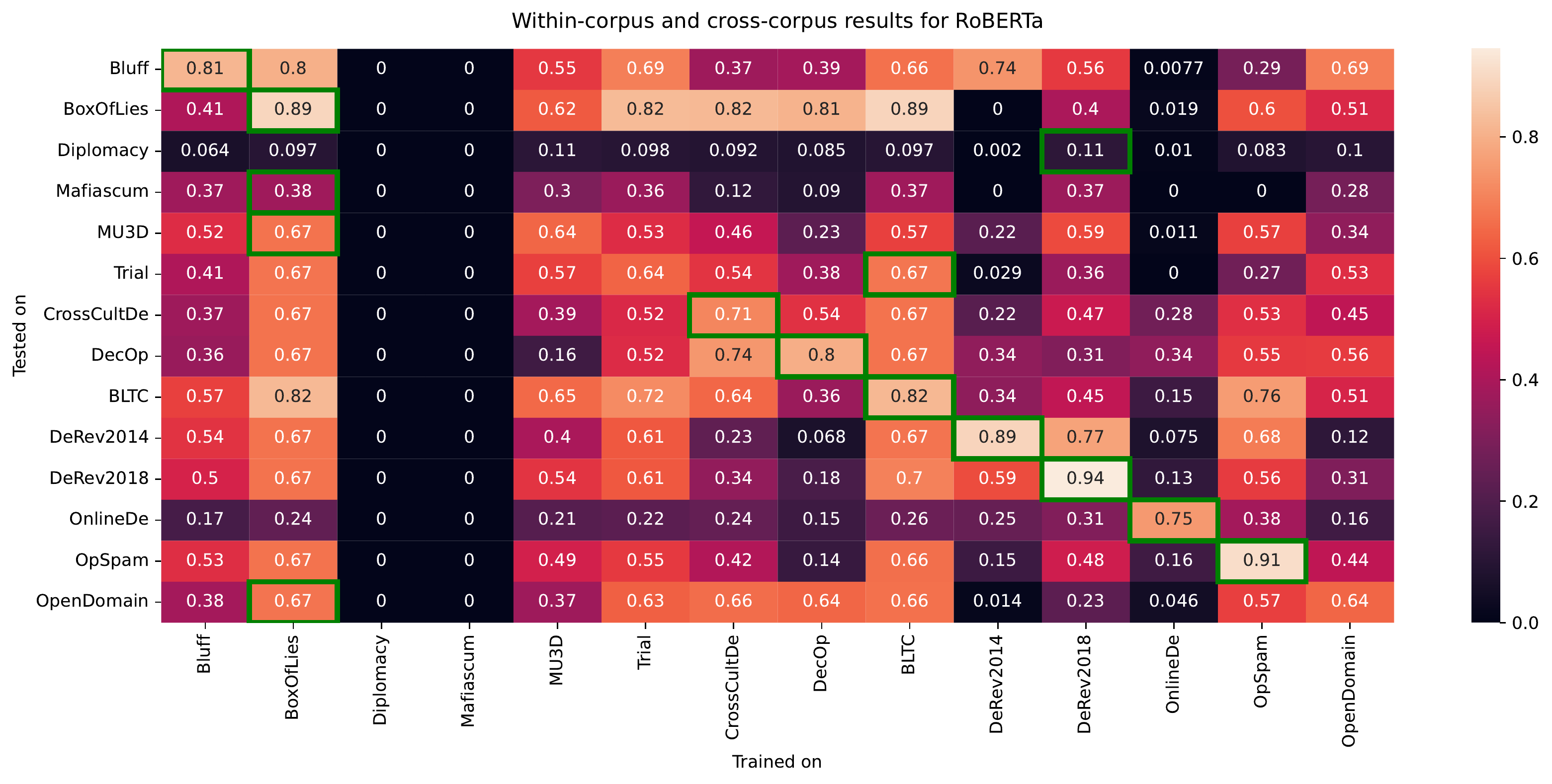}
  \caption{Performance of RoBERTa
    models with \F measure on the deception label. The best model on
    each test set is highlighted with a green box.}
  \label{fig:roberta-deception}
\end{figure*}

\section{Linguistic Correlation Analysis}
\label{lca}

To understand the generalizability of linguistic cues across different
dataset, we conduct a correlation analysis, similar to
previous studies that focused on isolated or smaller numbers of
corpora \citep{perez-rosas-mihalcea-2015-experiments,
  Skalicky2020PleasePJ}

\subsection{Method}
We aim at identifying frequently used features which are general
across domains. We build our analysis on the ``{Linguistic Inquiry and
  Word Count}'' \citep[LIWC22\footnote{\url{https://www.liwc.app/}},][]{pennebaker-2015-linguistic} and
Flesch-Kincaid \cite{kincaid1975derivation} and Gunning Fog
\cite{robert1952technique} readability scores as measures of
complexity or sophistication of language.\footnote{\url{https://pypi.org/project/readability/}.}

We use point-biserial
correlation\footnote{\url{https://docs.scipy.org/doc/scipy/reference/generated/scipy.stats.pointbiserialr.html}}
\citep{glass1996statistical} to measure the relation between deception
labels (discrete) and a score assigned by LIWC or readability
measurement (continuous).  The correlation value ranges from $-$1 to
$+$1.

\subsection{Results}

Table \ref{tab:linguistic} lists the features which show at least a
weak correlation ($>0.15$) with $p\leq 0.05$ for at least three datasets.
The positive and negative correlation values correspond to the strength of
association with truth and deception respectively.

Deceptive language is argued to have fewer self-references (``i'') and
more references to others (``shehe'', ``you''), as liars attempt to
distance themselves from their lies \cite{Newman2003,depaulo2003cues}.
Our analysis supports this hypothesis in the categories ``shehe'' and
``you'' for a substantial number of data sets.  Contrary to our
expectation, however, in 8 out of 14 datasets the category ``i'' is
seen to correlate with deception and not with truth, with an exception
of \crosscult ($\rho=.13$) and \derevb ($.39$).

Studies have attributed less cognitive complexity in language to
deceptive communication \cite{Newman2003, depaulo2003cues}.  Liars use
fewer words related to cognitive concepts (e.g., \textit{think},
\textit{believe}), which should correspond to a positive correlation
value for the category ``Cognition'' in LIWC. However, our analysis
corroborates this observation only in \bluff($\rho=.17$) and
\decop($\rho=.07$).

In general, we found no consistent linguistic cues across domains and
datasets in our analysis. This might be because deception is highly
sensitive to the goal of a lie and the stakes involved, which is not
consistent across the domains under consideration.

\section{Deception Detection Experiments}
The correlation analysis in the previous section showed that deception
cues do barely generalize across domains. This analysis might be
limited by the choice of categories, which motivates us to conduct
cross-corpus modeling experiments.

\subsection{Experimental Setup}
In the within-corpus setup, we fine-tune and evaluate Ro\textsc{bert}a
models \cite{liu2019roberta} on the same dataset via 10-fold
cross-validation. In the cross-corpus setting, we train on one corpus
and test on the other. To ensure comparability between these
experiments, we perform 10-fold cross-validation in both settings: we
also evaluate 10 times on the same corpus subsets in the cross-corpus
setup. This is not strictly required but ensures comparability.

We use the English Ro\textsc{bert}a-base, with 12 layers, 768 hidden-states, 12
heads and 125M parameters as available in the HuggingFace
implementation \citep{wolf-etal-2020-transformers}. We finetune with
default hyperparameters for 6 epochs using the Auto Model for Sequence
Classification.
\footnote{\url{https://huggingface.co/transformers/v3.0.2/model_doc/auto.html}}

\subsection{Results}
The heatmap in Figure \ref{fig:roberta-deception} shows the results as
\F measure for the deception label (Appendix \ref{append:results}
shows results for both labels). The diagonal corresponds to
within-corpus experiments.
For most datasets, the model shows better performance in the
within-corpus setting than in the cross-corpus evaluation. This is not
the case for \mud, \trial, and \opendomain, but the
difference is negligible ($0.04$).

Models on datasets from the same domain or which are otherwise similar
(\S~\ref{ss}) show comparably better results in the cross-corpus
setting.  For instance, training on \opspam and testing on \bltc
achieves an \F score of $0.76$ on the deception label. Training on \bltc and
testing on \opspam is however not as good ($0.66$).
Similar observations can be made for \dereva and \derevb, and
\crosscult and \decop.

The heatmap shows the lowest performance for \mafiascum and
\diplomacy, with an $\text{F}_1$=0. We assume that this is a result of
the imbalanced label distribution in \diplomacy and the long documents
in \mafiascum (see Table \ref{tab:distribution}). Similarly, the 
exceptionally good results on the \bol test set are due to the bias
towards the deceptive label (see appendix for \F score on truth label). 

Note that previous work reported other evaluation measures 
than \F, which makes this dramatically low performance difficult 
to compare. Our evaluation with accuracy (shown in the appendix in
Figure~\ref{fig:roberta-deception-acc}) appears to be more positive
with $.77$ and $.95$.

From the sub-par results on cross-corpus experiments, we conclude 
that generalization across domains and dissimilar datasets is 
challenging, even with pre-trained language models with rich contextual
information. In our future work, we plan to use this dataset to 
train models that can capture domain-independent cues of deception,
which can presumably generalize better across datasets.  

\section{Conclusion \& Future Work}

Different scientific disciplines have contributed to the creation of
deception datasets for textual communication in a variety of
domains. In this study, we present a comprehensive survey of deception
datasets in English available for research and compile
them into a unified deception dataset. We are not aware of any
previous work that considered a comparably large amount of corpora and
evaluated models between all of them. Some of the evaluation results
are encouraging, but particularly between dissimilar domains, the
generalization is limited and requires future research.

The RoBERTa-based classification experiments and
linguistic correlation analysis of deception cues demonstrate that it
is indeed challenging to generalize the concept of deception across
datasets, or domains. In the classification experiment results, the wildly 
diverging \F scores can be attributed to the complexity of the task as well as
to the limitations of the approach employed. In future work, we plan to
explore the reasons for this variability across datasets further.

Additionally, we acknowledge the need to address the issue of biased 
models, such as the ones trained on \mafiascum, \onlinede, and 
\diplomacy, which tends to favor truthful labels owing to the label imbalance
in these datasets, resulting in an \F score of 0. To overcome this 
challenge, we could employ techniques like oversampling to 
rectify the class imbalance and improve the reliability and 
effectiveness of our approach.

The goal of our future work is to create robust deception detection models 
that work reliably across corpora and domains. This includes understanding 
differences in the concept as it represents itself in these data and 
understanding differences in linguistic realization.  

Our \corpusname dataset 
serves as a valuable resource for future research enabling 
standardized data comparison, transfer
learning, and domain adaptation experiments.

\section*{Acknowledgments}
We thank Kai Sassenberg for fruitful discussions
regarding the concept of deception. Roman Klinger's research 
is partially funded by the German Research
 Council (DFG), projects KL 2869/1-2 and KL 2869/5-1.

\section*{Limitations}

The goal of the current study was to unify the resources available for
deception and report observations on cross-corpus and within-corpus
analyses. While reporting the baseline performance using RoBERTa, we
did not perform any optimization specific to the datasets. Hence,
better results might be reported in the papers which handle the
datasets or domains in isolation.

\section*{Ethical Considerations}
The datasets used in this research are publicly available
resources from previous studies. We have taken appropriate steps 
to ensure that we do not violate any license terms or intellectual 
property rights. Also, proper attribution is given to the 
original sources of the data. Deception is a sensitive topic, and
non-anonymous data should not be used. To the best of our
knowledge, all data sets that we considered have been compiled or
collected according to such standards.

The performance of deception detection systems is not perfect, 
making them unsuitable for examining the utterances of individuals
due to the threat of incorrect predictions. Even if automatic systems
might reach a close-to-perfect performance, we consider their
practical application to analyze and profile people unethical. However, there
might be use cases, for instance in forensics, that can be considered
ethical from a utilitaristic perspective.

Given the ethical implications of employing automated deception 
detection systems on individual, non-anonymous statements, 
we propose utilizing the resources collected and models developed 
on anonymous data. Any data analysis that could lead back to its 
origin must only be conducted with the data creator's informed 
consent and knowledge of potential consequences.

We consider the research in this paper to be fundamental, with the
goal of better understanding human communication.

\bibliography{anthology,custom}

\clearpage

\onecolumn

\appendix

\section*{Appendix}
\section{Details on the Aggregated Dataset}
\label{append:aggregation}

All datasets included in the unified collection contains one binary
label indicating whether an instance is truthful or deceptive, the
naming convention for which has been normalized retaining the original
label for backward compatibility. However, some datasets like
\citet{salvetti-etal-2016-tangled} and \citet{peskov-etal-2020-takes}
include an additional dimension for deception, where the former
differentiates between lying about a known object and lying about an
unknown object, and the latter contains annotations on the
\textit{perceived truthfulness} of the statement in addition to the
actual intention. For providing a unified format, we map both these
deceptive instances in \citet{salvetti-etal-2016-tangled} to one label
and since the perceived truthfulness is an independent annotation, we
do not take this label into account.

In addition to truth labels, datasets also contain additional
annotations like demographic information related to the author,
sentiment, valency of the instance and perceived truthfulness. We
retain only those annotation dimensions which are available for more
than one dataset which are age, gender, country, and sentiment

The unified dataset includes corpora that are available for research
purposes which are downloadable from source, made available directly
by the creators, or obtained from a consortium like the Linguistic
Data Consortium. We provide a script to automatically download all
datasets if they are available for download, which otherwise provides
instructions on how to obtain them. Once all datasets are populated in
their respective folders, a second script is used to generate the
unified dataset in json format. You can find the repository with
instructions to obtain the aggregated \corpusname, Unified Deception
Corpus, at  \url{https://www.ims.uni-stuttgart.de/data/unidecor}.
 
The following entry shows an example instance from the corpus.
\begin{lstlisting}[language=json,firstnumber=1,breaklines=true]
{
	"source": "OPEN_DOMAIN",
	"text_ID": "119_f_t_1",
	"text": "Thad cochran has been in the us senate since before the internet was
	invented.",
	"participant_ID": "NA",
	"age": "20",
	"sentiment": "NA",
	"language": "EN",
	"gender": "Female",
	"country": "United States",
	"original_label": "truth",
	"truth_label": "T",
	"topic_name": "statement",
	"domain": "opinion",
	"mode": "written",
	"split": null,
	"fold": null
}

\end{lstlisting}

\clearpage

\section{Additional Experimental Results}
\label{append:results}
\begingroup \centering
\includegraphics[scale=0.5]{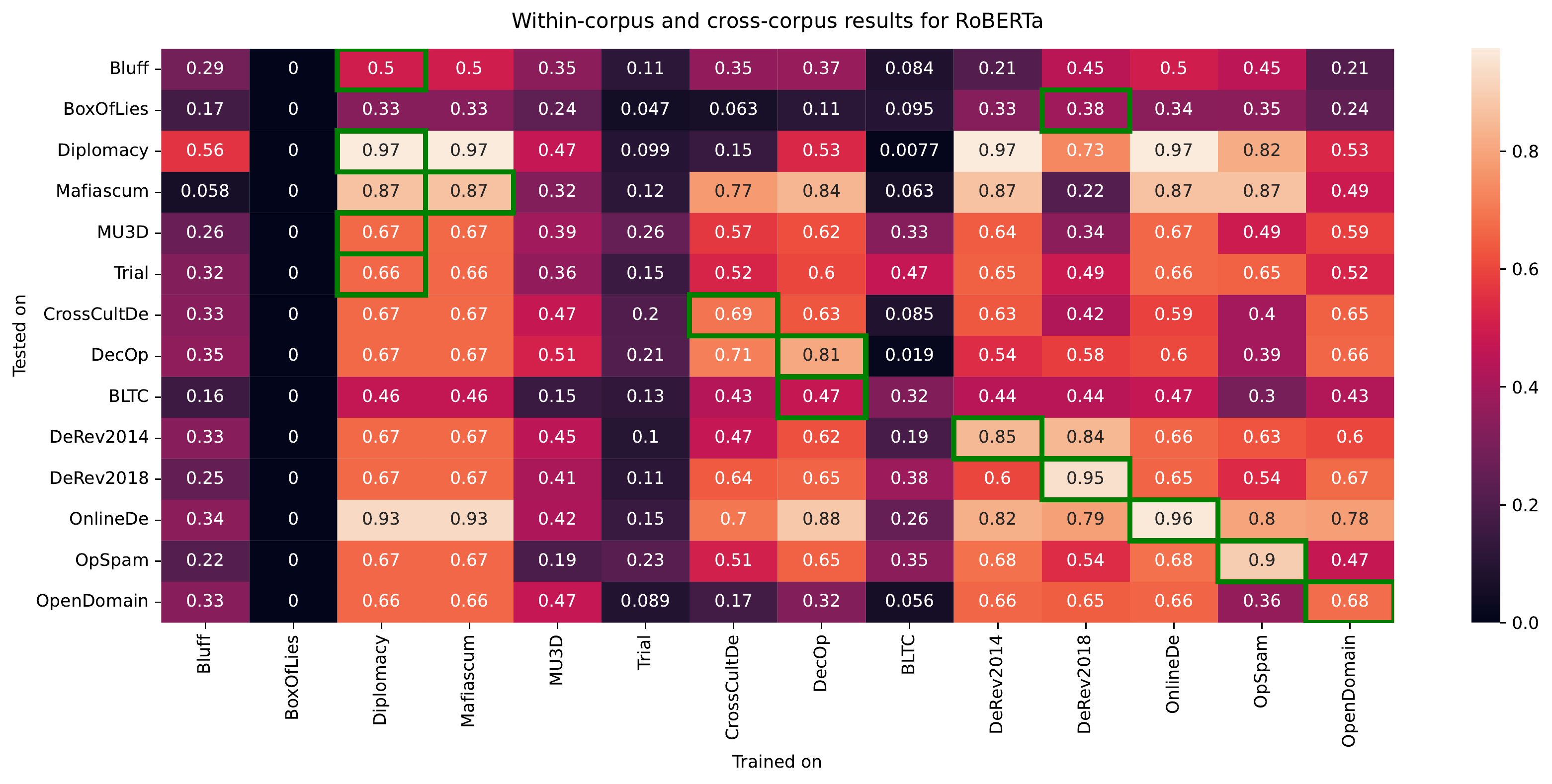}
\captionof{figure}{\label{fig:roberta-deception-truth}A heatmap
  representing the performance of RoBERTa model with the \F measure on
  the truth label across different
  datasets. Figure~\ref{fig:roberta-deception} in the main paper
  analogously shows the results for the deception category.}
\vspace{2cm}
\includegraphics[scale=0.5]{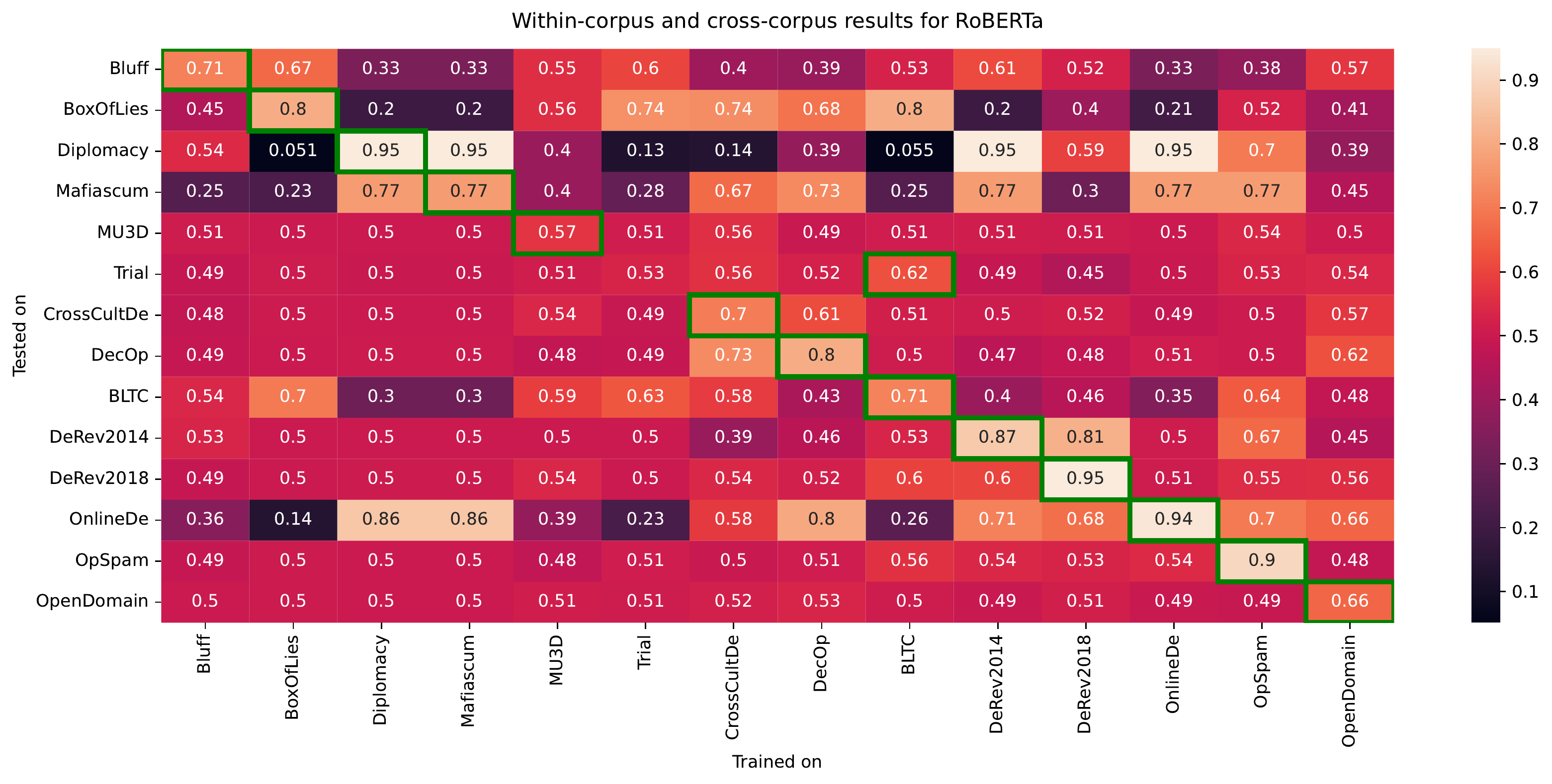}
\captionof{figure}{\label{fig:roberta-deception-acc}A heatmap
  representing the accuracy of RoBERTa model different datasets. As
  the categories of truth and deception and mutual exclusive in all
  our datasets, this corresponds to a micro-average of the results
  shown in Figure~\ref{fig:roberta-deception}
  and~\ref{fig:roberta-deception-truth}.}

\endgroup

\end{document}